\documentclass{article}
\usepackage{ijcai25}

\usepackage{times}
\usepackage{soul}
\usepackage{url}
\usepackage[hidelinks]{hyperref}
\usepackage[utf8]{inputenc}
\usepackage[small]{caption}
\usepackage{graphicx}
\usepackage{amsmath}
\usepackage{amsthm}
\usepackage{booktabs}
\usepackage{algorithm}
\usepackage{algorithmic}
\usepackage[switch]{lineno}
\usepackage{natbib}
\usepackage{amsfonts}
\usepackage{multirow}
\usepackage{subfig}

\title{ASCENT-ViT: \underline{A}ttention-based \underline{S}cale-aware \underline{C}oncept Learning Framework for \underline{E}nhanced Alignme\underline{nt} in Vision Transformers}

\author{
Sanchit Sinha$^1$
\and
Guangzhi Xiong$^1$\And
Aidong Zhang$^1$\\
\affiliations
$^1$University of Virginia\\
\emails
\{sanchit, hhu4zu, aidong\}@virginia.edu,
}

\begin{document}

\maketitle

\begin{abstract}
As Vision Transformers (ViTs) are increasingly adopted in sensitive vision applications, there is a growing demand for improved interpretability. This has led to efforts to \textit{forward-align} these models with carefully annotated abstract, human-understandable semantic entities - \textbf{concepts}. Concepts provide \textit{global rationales} to the model predictions and can be quickly understood/intervened on by domain experts. Most current research focuses on designing model-agnostic, plug-and-play generic concept-based explainability modules that do not incorporate the inner workings of foundation models (e.g., inductive biases, scale invariance, etc.) during training. To alleviate this issue for ViTs, in this paper, we propose \textbf{ASCENT-ViT}, an attention-based, concept learning framework that effectively composes scale and position-aware representations from multiscale feature pyramids and ViT patch representations, respectively. Further, these representations are aligned with concept annotations through attention matrices - which incorporate spatial and global (semantic) concepts. ASCENT-ViT can be utilized as a classification head on top of standard ViT backbones for improved predictive performance and accurate and robust concept explanations as demonstrated on five datasets, including three widely used benchmarks (CUB, Pascal APY, Concept-MNIST) and 2 real-world datasets (AWA2, KITS). 
The code can be found at: \url{https://anonymous.4open.science/r/sact-attention-AF4E/}

\end{abstract}

\section{Introduction}
\label{sec:intro}


With the surge of advanced deep learning (DL) methods, deep neural networks (DNNs), especially Transformer-based networks \citep{vaswani2017attention}, have been widely deployed in human communities for the benefit of our life \citep{dargan2020survey,shorten2021deep}. Research in developing high-performing, generalizable, and efficient architectures has led to experts classifying a family of fundamental architectures as `Foundation Models' \citep{bommasani2021opportunities}. In computer vision, the most widely utilized foundation model, Vision Transformers (ViT), shows great scalability and impressive performance in various downstream tasks, which promotes its use in real-world applications like autonomous driving \citep{ando2023rangevit,dong2021image}. However, with the development and application of such advanced DNNs, there is a growing concern about their lack of interpretability leading to hesitation in applying the technology in high-stakes areas such as medical diagnosis, facial recognition, finance, etc. Given the current situation, several studies on the interpretability of DNNs have emerged \citep{li2022interpretable}.


Current research on DNN interpretability can be broadly categorized into two classes - dubbed \textbf{backward} and \textbf{forward} alignment \citep{gabriel2020artificial}. \textit{Backward} alignment is post-hoc interpretation, which explains already trained models. Specifically, some post-hoc methods assign relative importance scores to different features considered \citep{ribeiro2016should,sundararajan2017axiomatic}, while some other approaches will rank the input training samples according to their importance to prediction \citep{koh2017understanding,ghorbani2019data}. On the other hand, \textit{forward} alignment methods try to design and train intrinsically interpretable models. One such approach attempts to incorporate human-understandable concepts into model architecture and training processes. Concepts can be considered as shared abstract entities across multiple instances, which provide a general understanding of the task processed by the model. Many studies have explored how to integrate the model's decision-making process with task-relevant concepts to achieve transparent and interpretable model predictions \citep{koh2020concept,pedapati2020learning,jeyakumar2020can,heskes2020causal,o2020generative,kim2018interpretability,li2018deep,alvarez2018towards,chen2019looks}. Formally, given a set of concept representations $C$ associated with an input sample $x$, the task prediction is $y = g(C)$, where $g$ is any interpretable function used to aggregate concepts. Multiple works such as \citep{alvarez2018towards} and \citep{agarwal2021neural} have postulated using a weighted linear sum of concepts as $g$. Other works such as \citep{koh2020concept} have postulated using a shallow feed-forward network as mapping from concepts to prediction.

With the success of attention mechanism \citep{vaswani2017attention} in designing large-scale networks for language \citep{devlin2018bert} and vision applications \citep{dosovitskiy2020image}, multiple diverse use-cases for attention have been proposed. A particular use case of attention in concept-based explainability is utilizing the attention mechanism in modeling the concept aggregation function $g$ as first proposed in \citep{rigotti2021attention}. Note that concepts are independent entities shared among different training samples, hence attention can be reliably used to assign ``relevance'' scores to the concepts relative to a prediction. 


ViTs \citep{dosovitskiy2020image} demonstrate significant improvement over traditional models like CNNs for large-scale vision applications. ViTs borrow their architecture from transformer-based language approaches \citep{vaswani2017attention,devlin2018bert}. A ViT first segments an image into equal patches, which are projected into an embedding space with positional information. The representation is then passed through stacked transformer blocks which is used in downstream tasks. ViTs have demonstrated better performance than CNNs, especially when pre-trained with large-scale data. However, due to the very architecture of ViTs being borrowed from language paradigm, they lack vision-specific \textit{inductive biases} like CNNs. CNNs have been specifically designed for image modalities, and capture explicit inductive biases such as - scale invariance and transformation equivariance. As an example, ViTs are demonstrated not to be scale-insensitive \citep{xu2021vitae}, a known strength of CNNs. Researchers have attempted to improve ViTs by incorporating CNN-like inductive biases in their architectures with great success \citep{liu2021swin,graham2021levit}. 

However, concept-based explainability approaches are still designed to be model-agnostic \citep{koh2020concept} and plug-and-play \citep{rigotti2021attention} to fit all possible use cases. As incorporating various inductive biases in the model architecture provides significant performance improvement, a detailed design decision should be made for concept-based explainability modules as well. For instance, explainability modules for ViTs can benefit from scale awareness in addition to patch awareness during concept learning in ViTs. As a consequence, we design a novel framework for ViTs that incorporates desirable properties (inductive biases) of ViTs - patch awareness and CNNs - scale invariance.
Specifically, in this paper, we propose \textbf{Attention-based Scale-aware Concept Learning Framework for
Enhanced Alignment in Vision Transformers (ASCENT-ViT)}, a concept-based explainability module for ViTs to effectively \textit{align} human-annotated concepts with model representations. Unlike model-agnostic plug-and-play explainability, ASCENT-ViT effectively composes inductive biases of CNNs and ViTs - scale invariance and image-patch relationship awareness respectively into scale and patch-aware representations. More precisely, ASCENT-ViT consists of 3 distinct modules - (i) \textbf{Multi-scale Encoding (MSE) Module}, which models concepts at various scales, (ii) \textbf{Deformable Multi-Scale Fusion (DMSF) Module}, which composes multi-scale concepts with patch embeddings, and (iii) \textbf{Concept-Representation Alignment Module (CRAM)}, which aligns concepts with the learned model representations. To summarize, our contributions are:
\begin{itemize}
    \item We propose ASCENT-ViT, a concept-based explainability method that effectively composes inductive biases from CNNs and ViTs to perform better concept-representation alignment. We are the first work to effectively integrate these inductive biases in concept-based explainability modules.
    \item We demonstrate through quantitative experiments that ASCENT-ViT improves both prediction accuracy and concept learning as compared to model-agnostic attention-based concept explainability modules across 7 different vision-transformer architectures.  
    \item We demonstrate through diverse real-world visual examples that ASCENT-ViT captures concept annotations missed by model-agnostic approaches and is robust to transformations and perturbations.
\end{itemize}

\section{Related Work}
\label{sec:related}

\noindent\textbf{Vision Transformers (ViTs) and Inductive biases.}
Multiple approaches have improved the original ViT \citep{dosovitskiy2020image} performance \citep{atito2021sit,gong2021ast}, efficiency \citep{chen2022vision} and explainability \citep{xu2023attribution}. Even though ViTs offer better downstream task performance as compared to CNN-centric approaches \citep{tuli2021convolutional, raghu2021vision}, the consensus among researchers remains that ViTs have much less stringent inductive biases and some approaches improve them \citep{xu2021vitae}. To alleviate this problem, namely - lack of scale invariance and transform equivariance, DETR \citep{zhu2020deformable} proposes an attention mechanism to incorporate scale features while SWIN \citep{liu2021swin} uses a shifted window approach in ViTs with great success. A few approaches have tried to improve ViT architectures with scale-awareness \citep{lin2023scale,guan2024enhanced}. 

\noindent\textbf{Concept-based Explainability and Attention.}
Concept-based explanations explain models using human-understandable concepts \citep{koh2020concept,jeyakumar2020can,kim2018interpretability,sinha2023understanding,sinha2024colidr}. Concepts are high-level attributes that are shared across different instances \citep{chen2019looks,koh2020concept}. Various explorations uncover the decision-making of black-box neural networks via concepts, including learning latent concept scores \citep{alvarez2018towards}, finding concept prototypes \citep{li2018deep,chen2019looks}, and activations \citep{kim2018interpretability,sinha2024self}.

\noindent\textbf{Comparison with related approaches.}
Recent approaches incorporate an external Attention matrix to align model representations with human-understandable concepts. Two related works to our method ASCENT-ViT are - Concept Bottleneck Models (CBMs) \citep{koh2020concept}, CT \citep{rigotti2021attention}, and BotCL \citep{wang2023learning}. Both CBMs and BotCL utilize an intermediate layer to perform alignment between concepts and representations but neither utilize spatial concept annotation maps - CBMs discretize the concepts to binary labels while BotCL is completely unsupervised and utilizes slot attention to learn representative concepts. For completeness, we use CBM as one of the baselines to compare our method's performance. The closest work to ASCENT-ViT is CT \citep{rigotti2021attention}, which utilizes a model-agnostic plug-and-play module to align concepts with representations. However, there are significant differences between our approach and CT. Firstly, their module is \underline{model-agnostic} and does not account for the inductive biases. Secondly, their method does not incorporate the effect of scale - making it \underline{fragile} to minor transformations. Finally, their method does not provide any analysis on its effectiveness over ViT \underline{scale and architecture} - a very important determinant of the extent of misalignment.

\section{Methodology}
\label{sec:method}

\begin{figure}[h]
    \centering
    \includegraphics[width=0.5\textwidth]{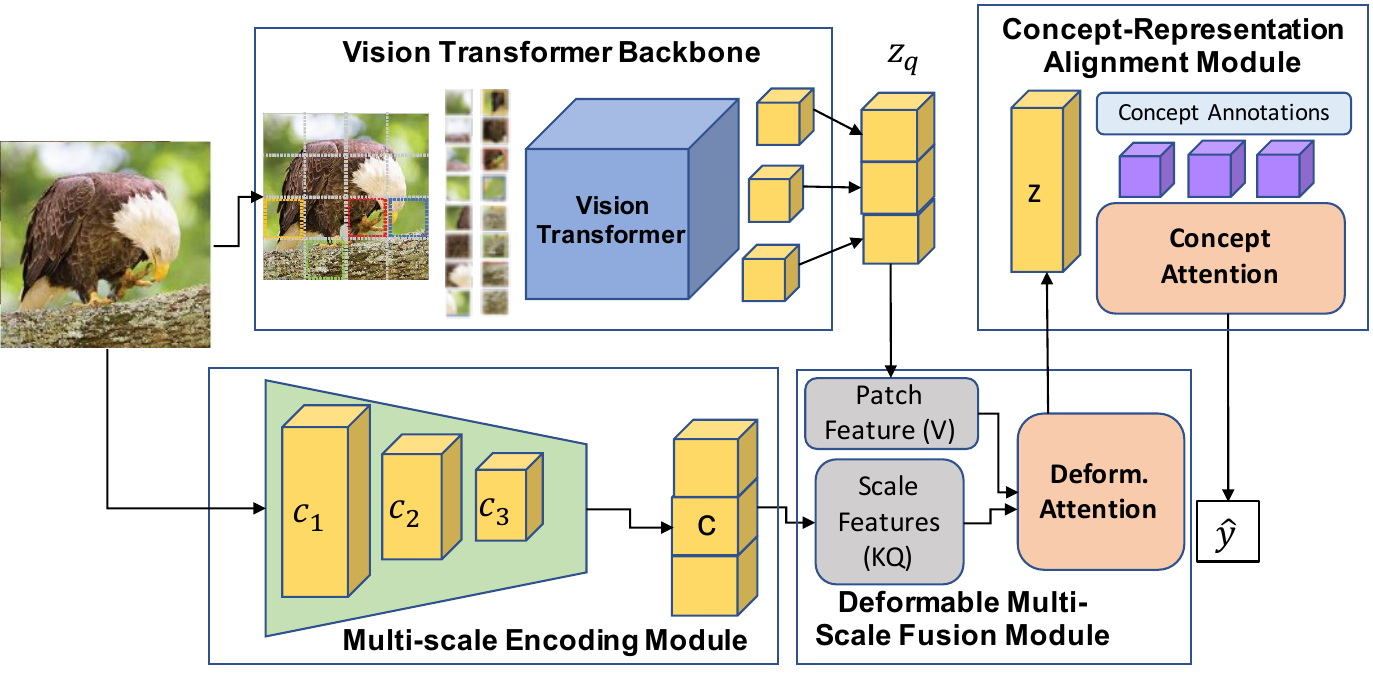}
    \caption{Schematic overview of the proposed ASCENT-ViT module. Given an input image, the Multi-scale encoding (MSE) module encodes representations at various scales in \textbf{c}. Patch-aware representations \(\mathbf{z_q}\) from the ViT are composed using the Deformable Multi-Scale Fusion (DMSF) Module. The Concept-Representation Alignment Module (CRAM) aligns the learned representations \(\mathbf{z}\) with concept annotations \(C\). \(\hat{y}\) is the model estimation of the true task label \(y\).}
    \label{fig:schematic}
    \vskip -15pt
\end{figure}

In this section, we introduce our proposed \textit{Attention-based Scale-aware Concept Learning Framework for
Enhanced Alignment in Vision Transformers (ASCENT-ViT)}. ASCENT-ViT is composed of three independent modules, namely - the Multi-scale encoding (MSE) module which extracts multi-scale features for enhanced scale awareness, the Deformable Multi-Scale Fusion (DMSF) Module which incorporates image scales using Deformable Attention, and the Concept-Representation Alignment Module (CRAM) which aligns scale and patch aware features with concept annotations. The ASCENT-ViT can be utilized as a classification head on top of standard ViT backbones for improved concept-based explanations. The overall schematic representation of the ASCENT-ViT is shown in Figure~\ref{fig:schematic}.

\noindent \textbf{Problem Setting:} A concept-based model consists of networks $f$ and $g$. For a given training sample $\{(x_i,y_i,C_i)\}$, where $x_{i} \in \mathbb{R}^{D}$ denotes the $i$-th training sample, $y_{i} \in \mathbb{R}$ is the target classification label for sample $x_i$, and $C_{i} \in \mathbb{R}^{T}$ is a vector of $T$ human annotated concepts, the network $f$ learns a mapping from image space to the concept space, i.e., $f: \mathbb{R}^{D} \rightarrow \mathbb{R}^{T}$, while the network $g$ maps concepts to the prediction space, i.e.,  $g: \mathbb{R}^{T} \rightarrow \mathbb{R}$. The training procedure entails supervising the concept space with human-annotated concepts and promoting the alignment of concepts and model representations. In our method, the concepts $C_i$ are the concept maps with pixel-level annotations.

\begin{figure}[h]
\centering
\subfloat[The Multi-scale encoding (MSE) Module.\label{fig:modules-sam}]{\includegraphics[width=0.85\linewidth]{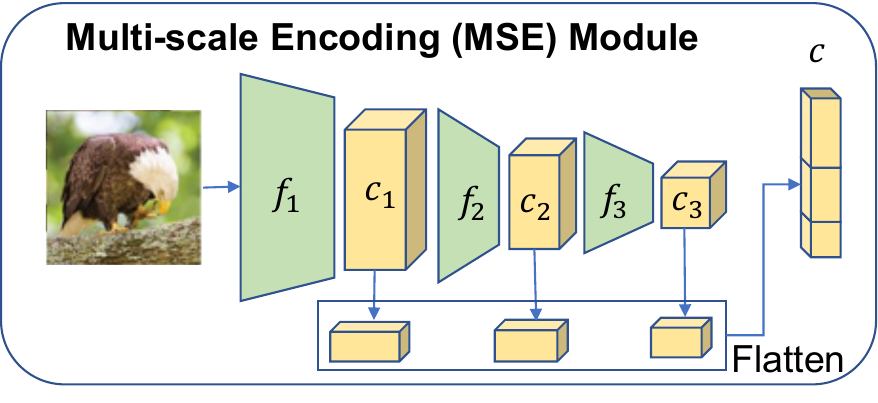}}\hfill
\subfloat[The Deformable Multi-Scale Fusion (DMSF) Module.\label{fig:modules-DMSF}]{\includegraphics[width=0.85\linewidth]{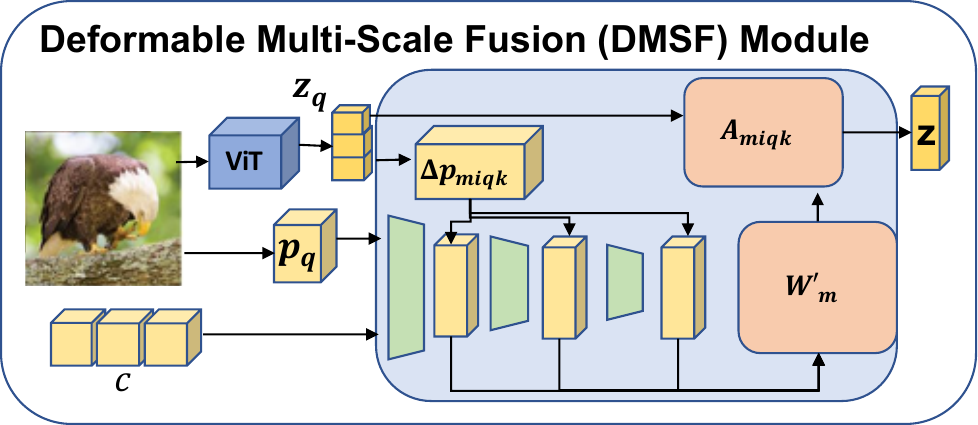}}\hfill

\caption{(a) Detailed view of the MSE module which extracts multi-scale features $\{c_i\}^S_1$ and concatenates them together into a vector $\mathbf{c}$. (b) Detailed view of the proposed DMSF module utilizing Deformable Attention operation to combine multi-scale features $\mathbf{c}$ with patch embeddings from ViT ($\mathbf{z_q}$) to output scale and patch-aware representations $\mathbf{z}$. }
\label{fig:modules}
\vskip -15pt
\end{figure}

\subsection{Multi-scale Encoding Module}
The Multi-scale encoding Module consists of a Convolutional Neural Network (CNN) $\mathbf{F}$ which can be thought of as ${S}$ stacked convolution blocks (with implicit activation and pooling operations) $f_i$, such that $\mathbf{F}= f_S \circ f_{S-1} \circ f_{S-2} ....,\circ f_{1}$. CNNs embody inductive biases in the form of convolution operations which capture the \textit{local semantics} of images at various scales, i.e., each block computes feature maps representing various scales. As a consequence, for a given input image $x \in R^{D}$, we extract ${S}$ multiscale features from various blocks of $\mathbf{F}$. Mathematically, each feature for a particular scale can be computed as
\begin{equation}
    c_i = f_i \circ f_{i-1}.. \circ f_1 (x), ~ i \in \{1,\cdots,S\}
\end{equation}
Subsequently, the combined multiscale features are concatenated by the MSE module to output a combined and spatially aware multi-scale feature $\mathbf{c}$. Mathematically, the scale-aware feature $\mathbf{c}$ is given as,
\begin{equation}
    \mathbf{c} = \{\text{flatten}(c_1),\text{flatten}(c_2),..\text{flatten}(c_S)\} 
\end{equation}
Note that the feature $\mathbf{c}$ is sometimes called the feature pyramid \citep{lin2017feature,chen2022vision} because it intuitively represents multiple scales of the same image, each scale progressively smaller than the previous scale. The architecture of the MSE module is presented in Fig. \ref{fig:modules} (a).

\subsection{Deformable Multi-Scale Fusion Module}
The DMSF module consists of two distinct steps. The first step computes an attention operation using deformed (scaled) outputs $\mathbf{c}$ from the MSE module with the Multi-Scale Deformable Attention (MSDA) operation \citep{zhu2020deformable} as discussed below. The second step composes a list of patch outputs $\mathbf{z_q}$ from a ViT backbone with multi-scale features computed from the MSDA operation using a combination of a learnable parameter ($\mathbf{I}$) and a tunable hyperparameter ($\psi$). 

\noindent\textbf{Multi-scale Deformable Attention (MSDA) Operation.}
As discussed before, the fundamental problem in the standard attention mechanism used in ViTs is the lack of scale and spatial awareness. To incorporate multi-scale feature maps, \cite{zhu2020deformable} proposed MSDA. The MSDA operation is characterized by a set of reference points (depicted by coordinates $\mathbf{p_q}$ and their offsets from a reference point $\Delta\mathbf{p}$) shared among features corresponding to multiple scales of the same image's feature maps which localizes semantics across scales. Mathematically, the MSDA operation for a set of feature-maps of various scales $\{c_i\}_1^S$ over a query-sample $\mathbf{z_q}$, the $MSDA(\mathbf{z_q}, \mathbf{p_q}, \mathbf{c})$ operation can be represented as,
\begin{equation}
    \sum_{m=1}^M \mathbf{W}_m [\sum_{i=1}^S\sum_{k=1}^K \mathbf{A_{miqk}} \cdot \mathbf{W'}_m c_i (\mathbf{\phi_i(p_q) + \Delta p_{miqk}}) ]
\end{equation}
where $M$ is the number of attention heads (indexed by $m$), $S$ is the number of scales (indexed by $i$), $K$ is the number of sampled reference points (indexed by $k$). $\mathbf{W}$ projects Value vector to multiple heads while $\mathbf{W'}$ is the inverse projection. $\mathbf{A}$ denotes the attention matrix, and $\Delta \mathbf{p_{miqk}}$ represents offset corresponding to sampled reference points $\mathbf{p_q}$, for each head and scale. Correspondingly, each element in $\mathbf{A}$ is referenced as head ($m$) and scale ($i$), while $q$ and $k$ represent the query and key indices, implying the number of reference points sampled is $S * K$. The function $\phi$ scales up/down the reference points to corresponding scales as $c_i$. Intuitively, the module can be understood as utilizing multi-scale features as key/value vectors that transform a patch-aware vector ($\mathbf{z_q}$) into a patch and scale-aware vector ($\mathbf{z}$).

\noindent\textbf{Adaptive Patch Composition.}
Note that the output from the ViT backbone is represented as $\mathbf{z_q}$ (Figure~\ref{fig:schematic}), which is treated as the query vector for the MSDA operation. Even though the MSDA operation introduces scale awareness in the outputs of the ViT backbone, a tradeoff between patch-specific semantics and scale awareness is desirable due to the wide gulf in the inductive biases in them. To alleviate this, we introduce a learnable vector $\mathbf{I}$ which controls the contribution of scale features, and a scalar hyperparameter $\psi$ which weighs the effect of the MSDA operation on the final patch vector. Mathematically, the final output of the DMSF module is as follows:

\begin{equation}
    \mathbf{z} = norm(\mathbf{z_q} + \psi \cdot \mathbf{I} \cdot MSDA(\mathbf{z_q}, \mathbf{p_q}, \mathbf{c}))
\end{equation}
where $norm$ represents the layer norm, $\mathbf{z_q}$ represents the input from the ViT backbone, $\mathbf{p_q}$ represents the reference points while $\mathbf{c} = \{c_i\}_1^S$ represents the multi-scale features computed by the MSE module. Note that adaptive composition is sensitive to the initialization of the $\mathbf{I}$ vector and the tunable hyperparameter $\psi$. An overview of the DMSF module is shown in Fig. \ref{fig:modules} (b).

\subsection{Concept-Representation Alignment Module}
\begin{figure}[h]
    \centering
    \includegraphics[width=0.4\textwidth]{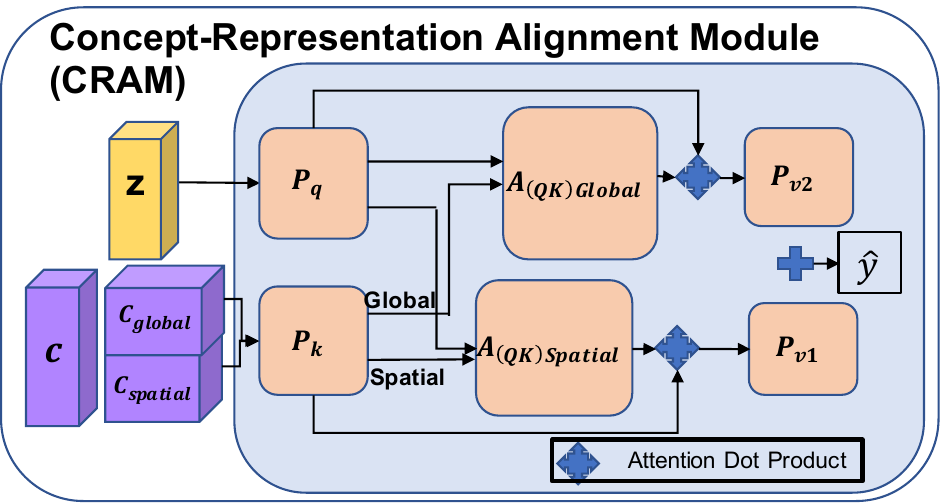}
    \caption{Detailed view of the Concept-Representation Alignment Module (CRAM). CRAM aligns scale and patch-aware representations ($\mathbf{z}$) and human-annotated concepts ($C$). The matrices $P_q$, $P_k$ and $P_{v1}$,$P_{v2}$ are projection matrices while $\mathbf{A_{Spatial}}$ and $\mathbf{A_{Global}}$ are attention matrices between concept projections and patch embedding projections. The final prediction $\hat{y}$ is the average of outputs between spatial and global attention operations. }
    \label{fig:ctc}
    \vskip -15pt
\end{figure}

The Concept-Representation Alignment Module (CRAM) is inspired by the Attention mechanism proposed in \citet{vaswani2017attention} wherein, an attention matrix $\mathbf{A_{att}}$ computation is conditioned on three vectors of the same dimensions - Query (Q), Key (K), and Value (V). Mathematically, the attention operation is as follows:
\begin{equation}
    \mathbf{Att}(Q,K,V) = \text{softmax}(\frac{QK^T}{\sqrt{dim}})V = \mathbf{A_{att}}V 
\end{equation}
The attention matrix attends to each value in the Query and Key vectors with a composite score. CRAM treats the Query vector as the patch representation \(\mathbf{z}\) from the DMSF module and the Key Vector as Human-annotated Concepts $C$. As patches and concepts are continuous and one-hot respectively, projection matrices convert them to continuous vectors. Mathematically, the attention matrix for patch embeddings ($\mathbf{z}$) and concepts ($C$) associated to an input sample $x$ is calculated as follows:
\begin{equation}
    \mathbf{A_{QK}} = \text{softmax}(\frac{(\mathbf{z}P_q)(CP_k)^T}{\sqrt{dim}})
\end{equation}
where the matrices $P_q$ and $P_k$ are learnable projection matrices transforming patches and concepts to Query and Key vectors respectively, while $dim$ represents the dimension of the patch and concept embeddings. Intuitively, each entry in the matrix $\mathbf{A_{QK}}$ encapsulates the contribution of a concept on a patch and is a measure of alignment. The final prediction is then calculated by multiplying the attention matrix with concept projections  ($V = CP_k$) followed by a final learnable projection matrix ($P_v$) to map to logits. Mathematically each logit $\hat{y}$ in the prediction $\hat{y}$ can be calculated as an average of concept contributions on each patch.
\begin{equation}
    \hat{y} = \frac{1}{|\mathbf{z}|}\sum^{|\mathbf{z}|} \mathbf{A_{QK}} V P_v
\end{equation}
where \(|\mathbf{z}|\) stands for the dimension of \(\mathbf{z}\). In practice, two different attention matrices ($\mathbf{A_{Spatial}}$ and $\mathbf{A_{Global}}$ - the $QK$ subscript is omitted for clarity) are calculated for different subsets of concepts - spatial and global. The spatial concepts can be localized, while the global concepts are more semantic and cannot be localized. The final output is then the sum of logits computed from both spatial and global attention matrices. Mathematically the sum can be written as:
\begin{equation}
    \hat{y} = \frac{1}{|\mathbf{z}|}\sum^{|\mathbf{z}|} \mathbf{A_{Spatial}} V  P_{v1} + \frac{1}{|\mathbf{z}|}\sum^{|\mathbf{z}|} \mathbf{A_{Global}} V  P_{v2}
\end{equation}
where $P_{v1}$ and $P_{v2}$ are different versions of $P_{v}$ associated with $\mathbf{A_{Spatial}}$ and $\mathbf{A_{Global}}$, respectively. The detailed description of CRAM is shown in Figure~\ref{fig:ctc}.

\subsection{End-to-end Training of ViT with ASCENT-ViT}
The overall schematic pipeline of our approach is depicted in Figure~\ref{fig:schematic}. As MSE, DMSF and CRAM modules are standalone modules that can be appended on top of any standard ViT backbone. To sum up, the inference step for an input image $x$ utilizing ASCENT-ViT computes the patch embeddings $\mathbf{z_q}$ through the following procedure: 
\begin{equation}\small
    \hat{y} = CRAM(\mathbf{z}, C),~~\text{where}~\mathbf{z} = DMSF(\mathbf{z_q},~ \mathbf{c}),  \mathbf{c} = MSE(x)
\end{equation}

\noindent \textbf{Training Objective.} 
The final training objective includes the task prediction objective utilizing a classification loss. For datasets with explicit localization of concepts in the image (e.g. CUB), the attention matrix $\mathbf{A_{Spatial}}$ can be regularized by human-annotated localizations. Mathematically, the training objective can be given as:
\begin{equation}
    \mathcal{L}_{ASCENT-ViT} = \mathcal{L}(\hat{y},y) + \lambda  || \mathbf{A_{Spatial}} - H ||_{F} 
\end{equation}
where $\mathcal{L}$ represents any classification loss, $H$ denotes human-annotated localization points for the concepts, $\lambda$ is a tunable hyperparameter and the subscript $F$ is the Frobenius norm.

\section{Experiments and Results}
\label{sec:results}

\subsection{Dataset Description}
    \noindent \textbf{(1) CUB200} \citep{wah2011caltech}: The Caltech-UCSD Birds-200-2011 dataset consists of 11,788 photos of 200 different classes of birds annotated with concepts representing physical traits of birds like wing color, beak size, etc.
    \noindent\textbf{(2) AWA2} \citep{xian2018zero}: Animals with Attributes-2 consists of 37,322 images of a combined 50 animal classes with 85 binary concepts like number of legs, presence of tail, etc. We utilize Segment Anything Model (v2)\footnote{\url{https://github.com/facebookresearch/sam2}} to annotate 10 selected concepts.
    \noindent\textbf{(3) KITS} \citep{heller2023kits21}: The Kidney Tumor Segmentation (KITS) challenge dataset consists of images of kidneys where the task is segmentation of renal cysts and tumors. We augment each image with the concepts from the meta-data, namely - kidney, tumor, and cysts. We sample 10000 images with an 80:20 train/test split.
    \noindent\textbf{(4) PascalAPY} \citep{farhadi2009describing}: We utilize Pascal Object Recognition Dataset as processed and described in \citep{rigotti2021attention}. After processing, it consists of 14350 images with 20 attributes each. 
    \noindent\textbf{(5) C-MNIST} \citep{sinha2023understanding}: Concept-MNIST entails augmenting MNIST dataset with two ``spatial'' concepts in the form of curved and straight lines. We consider two tasks - even/odd and digit classification.

\begin{table*}[t]
\centering
\resizebox{\textwidth}{!}{%
\begin{tabular}{|c|c|c|c|c|c|c||c|c|c|c|}
\hline
 &  & \multicolumn{2}{c|}{\textbf{CUB200}} & \multicolumn{2}{c|}{\textbf{AWA2}} & \multicolumn{1}{c||}{\textbf{KITS}} & \multicolumn{2}{c|}{\textbf{Concept-MNIST}} & \multicolumn{1}{c|}{\textbf{Pascal aPY}}  \\
 \hline
\textbf{Head} & \textbf{Backbone}  & Accuracy & Px. TPR & Accuracy &  Px. TPR  & Px. Acc.  & Class. & Odd/Even & Accuracy \\
\hline
CBM & ViT-Large & 81.03  & {-} & 53.14 & {-} & {-} & 93.83 & 97.12 & 81.1  \\
\hline
\multirow{3}{*}{CRAM (CT)} & ViT-Base & 84.41 $\pm$ 0.3 & 82.25 $\pm$ 0.3  & 62.3 $\pm$ 0.2 & 64.02 $\pm$ 0.5 & 91.89 $\pm$ 2.4  & 95.71 $\pm$ 0.1 & 97.58 $\pm$ 0.1 & 81.0 $\pm$ 0.8 \\
& ViT-Large   & 86.31 $\pm$ 0.2 & 83.55 $\pm$ 1.1 & 63.6 $\pm$ 0.1 & 68.04 $\pm$ 1.4 & 94.33 $\pm$ 2.0  & 95.74 $\pm$ 0.1 & 97.94 $\pm$ 0.1 & 81.3 $\pm$ 0.8 \\
& SWIN   & 85.71 $\pm$ 0.1 &  82.28 $\pm$ 0.8  & 62.3 $\pm$ 0.1 & 67.84 $\pm$ 0.7 & 92.81 $\pm$ 1.8  & 94.83 $\pm$ 0.1 & 97.68 $\pm$ 0.1 & 81.3 $\pm$ 0.8 \\
\hline
\multirow{3}{*}{ASCENT-ViT} & ViT-Base  & 85.71 $\pm$ 0.3 & 83.93 $\pm$ 0.2  & 62.3 $\pm$ 0.1 & 64.11 $\pm$ 1.2  &  92.33 $\pm$ 0.8 & \textbf{95.83 $\pm$ 0.1} & 97.91 $\pm$ 0.1 & 81.4 $\pm$ 0.8  \\
& ViT-Large   & \textbf{87.26 $\pm$ 0.3} & \textbf{87.21 $\pm$ 0.9} & \textbf{63.7 $\pm$ 0.1} & \textbf{73.12 $\pm$ 0.3}  & \textbf{95.84 $\pm$ 1.9} &  \textbf{95.83 $\pm$ 0.1} & \textbf{97.91 $\pm$ 0.1} & \textbf{81.9 $\pm$ 0.8} \\
& SWIN  & 85.74 $\pm$ 0.1 & 83.41 $\pm$ 1.3  & 62.5 $\pm$ 0.4 &  69.66 $\pm$ 0.4 & 93.02 $\pm$ 1.3  &   94.83 $\pm$ 0.1 & 97.91 $\pm$ 0.1 & 81.4 $\pm$ 0.8  \\
\hline
\end{tabular}}
\caption{We report Accuracy and True Positive Rate (TPR) for CUB200 and AWA2 datasets, over backbones ViT-Base, ViT-Large, and SWIN augmented with CRAM (Rows 2-4) and ASCENT-ViT augmented backbones (Rows 5-7). For the KITS dataset, we report the total pixel accuracy. For Concept-MNIST, we report accuracy on two tasks - Digit classification and Odd/Even detection; for Pascal aPY, we report both accuracy on object detection task. Refer to Appendix for details on task settings. All results are averaged over 3 seeds, and std dev is reported.}
\vskip -15pt
\label{tab:performance}
\end{table*}

\subsection{Baselines and Evaluation Metrics}
\label{sec:metrics}
\noindent \textbf{Comparision Baselines.}
As discussed in Related Work, our method is the first to utilize inductive biases in attention-based concept explainability. The closest comparison to our method is CBM \citep{koh2020concept} (CBMs do not utilize concept attention maps, only global concepts) and \citet{rigotti2021attention} which only utilizes an inferior attention-only architecture similar to only a CRAM module. \textbf{We denote \citep{rigotti2021attention} as ``Only CRAM''}. We compare our approach on a variety of ViT baselines - with ViT-Base, ViT-Large \citep{dosovitskiy2020image}, SWIN \citep{liu2021swin} for all datasets. ViT-Base and ViT-Large are chosen to demonstrate the effect of scale awareness, while SWIN is chosen to demonstrate the effect of inductive bias in the backbone. In addition, we also compare the CUB dataset on DeIT \citep{touvron2021training}, ViT-Base (Dino) \citep{caron2021emerging}. SWIN incorporates inductive biases from CNNs. Dino is a robustly trained ViT while DeIT is a distilled version, further capturing the diversity in training procedures. 

\noindent \textbf{Evaluation Metrics.} 
The evaluation metric for task prediction is the accuracy in the five datasets.
For datasets with spatial concept annotations, namely CUB, AWA2, and KITS, we utilize the Pixel True Positive Rate (Px. TPR) which measures the percentage of correctly identified concept annotations. Due to the extremely precise and sensitive setting of KITS dataset, we measure the total pixel accuracy which encompasses both accurately identified concept and nonconcept pixels. For datasets with no spatial concept annotations (only global) - namely C-MNIST and Pascal APY, we utilize the 0-1 error (misclassification) for concepts, as only binary concepts are considered. Additionally, for CUB200, as localization information for each concept is present, the concept errors are directly evaluated with the Frobenius norm of the difference between ground-truth annotations and predicted concept attention matrices (in the appendix).

\subsection{Implementation Details}
\noindent \textbf{ViT Backbone.} 
We set the patch size to correspond to 16x16 pixels. Each image is resized to (224,224), and hence a patch sequence is (224/16, 224/16) = (14,14). Following \citet{rigotti2021attention}, we append the \textless CLS\textgreater token to the patch sequence after positional embeddings encapsulating global semantics. This results in an input sequence of length of 196+1 = 197 tokens. The internal embedding size ($dim$) of ViT is 1024.

\noindent \textbf{MSE Module.} We utilize three scales in the Multiscale Encoding Module, i.e., $S=3$. We utilize scales 1/8, 1/16, and 1/32 corresponding to scale-aware vector $\mathbf{c}$ composed of sizes: $c_1=1024$, $c_2=196$, and $c_3=16$, making the size of $\mathbf{c} = 1029$. The first convolution block consists of three convolution layers with batch norm followed by Max pooling. The following convolution blocks consist of a single convolution layer of kernel size=3 and stride=2 with batch norm.  

\noindent \textbf{DMSF Module.} We utilize Multiscale Deformable Attention \citep{zhu2020deformable}. We used 16 attention heads and four reference points, along with layer norms for each key, query, and value vector. We strip \textless CLS\textgreater token before passing through the DMSF module and append it to the output. The initialization value of $\mathbf{I}$ is set as 0.01. The value of $\psi$ is tunable and set as 1 for CUB, 2 for Concept-MNIST and Pascal aPY, and 0.5 for AWA2 and KITS.

\noindent \textbf{CRAM Module.} We utilize the same embedding dimensions, i.e., $dim=1024$ for Key, Query and Value vectors. The number of attention heads is set as 2 for both global and spatial concept attention matrices. (in Appendix)

\noindent \textbf{Training Details.} We train each dataset and backbone for 50 epochs with early stopping. The maximum learning rate is set at 5e-5 with a linear warmup for the first 10 epochs followed by Cosine decay. The batch size is 16 for each dataset with mixed precision (16-bit default) optimization. 

\subsection{Quantitative Results}
\subsubsection{Task Prediction Performance.}
Table~\ref{tab:performance} reports the task prediction performance across all five datasets with ViT backbones using CBM (Row-1), augmented with only CRAM (Rows 2-4) and ASCENT-ViT (Rows 5-7). Rows 2-4 list the prediction performance of backbones augmented with only the CRAM module on - ViT-Base, ViT-Large, and SWIN. Rows 5-7 list the performance on the same architectures using ASCENT-ViT. ASCENT-ViT outperforms CBMs and only CRAM on all datasets. Interestingly, on SWIN backbones the performance of ASCENT-ViT and `only CRAM' is on par - implying that SWIN captures scale inductive biases better than ViTs but performs worse than ViT-Large (a larger model). These results demonstrate that both size and inductive biases of ViTs are important factors in improving prediction and concept performances. Our module is effective in introducing scale awareness in standard ViTs and improving their performance.

\subsubsection{Concept Annotation Performance.}
As concept performance metrics are different for datasets with spatial concept annotations and global annotations, we discuss the results individually as reported in Table~\ref{tab:performance}.

\noindent\textbf{True Positive Rate (TPR):} For CUB200 and AWA2, we observe that pixel-wise TPR improves by $\sim$4\% on CUB200 and $\sim$5\% implying that ASCENT-ViT captures pixels of interest much better than only CRAM. In Fig~\ref{fig:explain-img}, we demonstrate the effect on random test images where ASCENT-ViT captures concept annotations more effectively than only CRAM.

\noindent\textbf{Pixel Accuracy:} For KITS, we report total pixel accuracy which includes both concept and background annotations. Note that due to the sensitive nature of KITS, it is important to correctly identify both relevant (cysts) and irrelevant concepts (kidney surface). We observe ASCENT-ViT outperforms only CRAM by $\sim$1.5\% over all architectures on KITS.


\noindent\textbf{Misclassification Errors} (Appendix):
For datasets with only global concepts and no spatial annotations, we report the misclassification error. We observe ASCENT-ViT outperforms only CRAM on both C-MNIST and Pascal aPY.

\begin{figure}[h]
  \subfloat[Effect of number of heads\label{fig:ablation-heads}]{\includegraphics[width=0.48\linewidth]{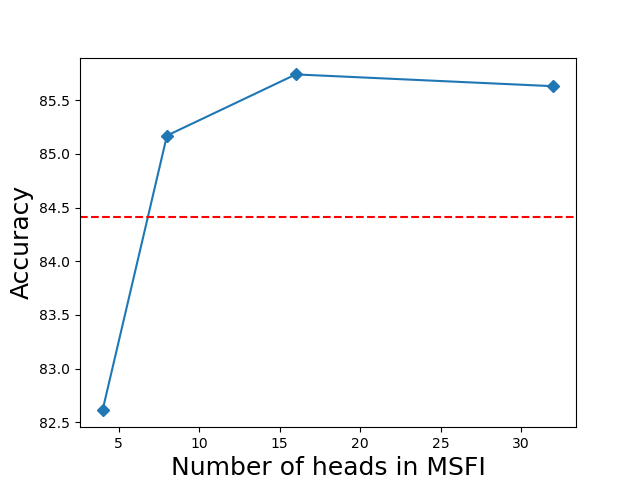}}
  \subfloat[Effect of $\psi$\label{fig:ablation-psi}]{\includegraphics[width=0.48\linewidth]{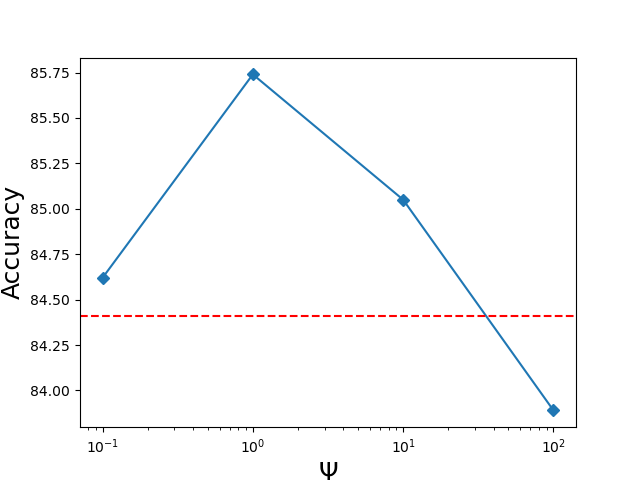}}
\vskip -5pt  
\caption{Effect of most relevant hyperparameters - attention heads and $\psi$ in the DMSF module on performance. The dotted red line shows the baseline performance of utilizing only CRAM.}
\vskip -10pt
\label{fig:ablation}
\end{figure}

\subsection{Ablation Studies} 
\label{sec:ablation}
\subsubsection{Hyperparameter Ablations.}
We demonstrate the effect of the most important hyperparameters in the DMSF module in Figure~\ref{fig:ablation}. For all experiments, a ViT-Base model is utilized and trained on the CUB dataset. Figure~\ref{fig:ablation-heads} demonstrates the effect of increasing the number of attention heads on the prediction performance. We observe that the increased number of attention heads improves performance by attending to various scales effectively. However, the maximum performance is observed at 16 heads implying that too many heads might result in the model focusing on non-relevant parts. Next, Figure~\ref{fig:ablation-psi} demonstrates the predictive performance as a function of increasing values of $\psi$ (i.e., increasing weight of scale-aware features on $\mathbf{z}$). Here, we observe that a tradeoff exists between $\psi$ and predictive performance. The reason for this behavior is with higher $\psi$, scale-aware inductive biases begin dominating the patch embeddings and the performance gravitates towards being similar to models with high CNN-like inductive biases (Table~\ref{tab:performance}).

\begin{table}[h]
\centering
\resizebox{0.42\textwidth}{!}{
\begin{tabular}{c|c|c|c|c}
\hline
  & \multicolumn{2}{c|}{\textbf{Concept-MNIST} } & \textbf{CUB200} & \textbf{AWA2} \\
\hline
\textbf{Modules} & Odd/Even & Class. & Class. & Class. \\
\hline
ViT-base & 97.12 & 93.82 & 82.91 & 61.7 \\
w/ CRAM  & 97.78 & 95.71 & 84.41 & 62.1 \\
w/ CRAM+MSE & 97.87  & 95.65  & 85.17 & 62.7\\
\textbf{ASCENT-ViT} & \textbf{97.88} &  \textbf{95.84} & \textbf{85.73} & \textbf{63.4}\\
\hline
\end{tabular}}
\caption{Task prediction performance ablating on proposed modules of ASCENT-ViT over a single run on tasks. NOTE: ASCENT-ViT implies Backbone w/ MSE+DMSF+CRAM.}
\vskip -10pt
\label{tab:ablation-modules}
\end{table}

\subsubsection{Module Ablations.}
Next, we progressively demonstrate each submodule's effect in Table~\ref{tab:ablation-modules}. Line-1 lists the performance of ViT-base with no concept information. Next, Line-2 lists the performance only using the CRAM module. Line-3 lists the performance with both the MSE and CRAM modules. As we observe, adding scale-aware features indeed improves performance. The last line (Line-5) lists performance when the DMSF module performs feature fusion - performing the best. 

\begin{figure}
    \centering
    \includegraphics[width=0.4\textwidth]{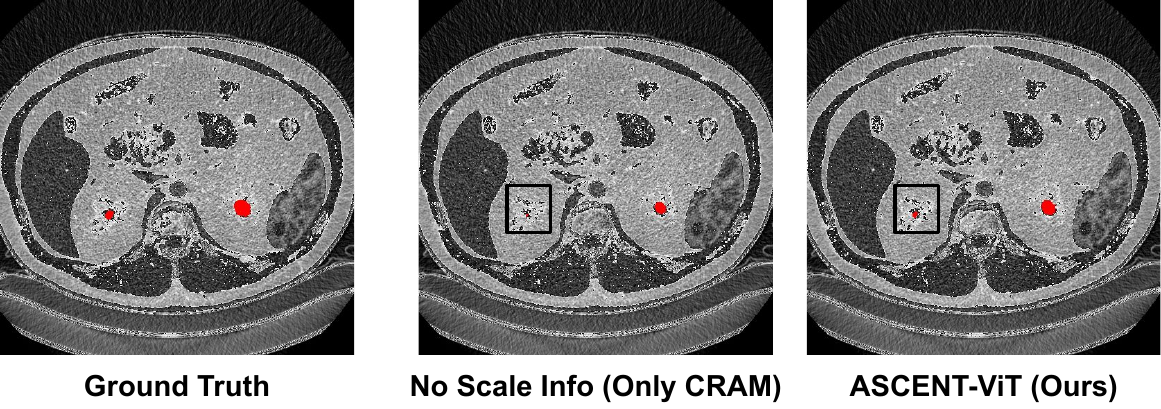}
    \caption{Effect of Scale: Visualized concept annotations for a correctly classified sample from KITS. As seen, ASCENT-ViT identifies \underline{small} `cyst' annotation much more accurately (Acc=91.2\%) than only CRAM (Acc=88.4\%) where no scale information is utilized.}
    \label{fig:kidney-CRAM}
    \vskip -10pt
\end{figure}

\begin{table}[h]
    \centering
    \resizebox{0.49\textwidth}{!}{
    \begin{tabular}{c|c|c|c}
        \hline
         \textbf{Model Name} & \textbf{Inductive Bias} & \textbf{Only CRAM} & \textbf{ASCENT-ViT} \\
         \hline
         ViT-Large & Positional (Pos.) & 86.31 $\pm$ 0.2  & \textbf{87.26 $\pm$ 0.3}  \\
         DeIT-Base & Positional (Pos.) & 76.81 $\pm$ 0.8 & \textbf{76.96 $\pm$ 0.3}  \\
         SWIN (1x1) & CNN-like & 86.44 $\pm$ 0.8 & \textbf{ 86.72 $\pm$ 0.6} \\
         ViT-Base (Dino) & Pos. + Invariance & 76.5 $\pm$ 0.2 & \textbf{77.2 $\pm$ 0.1} \\
         \hline
    \end{tabular}}
    \caption{Performance on various ViTs on CUB200 using only CRAM and ASCENT-ViT. We observe that architectures with CNN-like inductive biases show smaller improvements and have comparable performance for both ASCENT-ViT and CRAM.}
    \vskip -15pt
    \label{tab:model-type-effect}
\end{table}

\subsubsection{ViT Backbone Ablations.} Table~\ref{tab:model-type-effect} lists the prediction performance of DeIT-Base, SWIN 1x1, and ViT-Base (Dino), variants on CUB200 averaged over three seeds. We observe that ASCENT-ViT outperforms CRAM on models with either a missing Positional or CNN-like inductive bias (DeIT, SWIN, ViT-Base (Dino)). This observation is expected as ASCENT-ViT provides explainability which combines both Positional and CNN-like scale-invariant inductive biases. Note that on SWIN, which contains pre-encoded CNN-like inductive biases, ASCENT-ViT is on par with CRAM. This observation reaffirms the motivation of our approach - incorporating multiple inductive biases in the explainability module provides better explanations. We compare the quality of explanations for random test set images across ViT architectures in Appendix. 

\noindent\textbf{Overhead of Explanations:} We also provide the overhead of the additional modules as a percentage of model parameters (Appendix). In most cases, the models augmented with CRAM consist of $\simeq$2\% additional parameters than just the backbone implying a negligible increase in the training times.

\begin{figure}[h]
    \centering
    \includegraphics[width=0.43\textwidth]{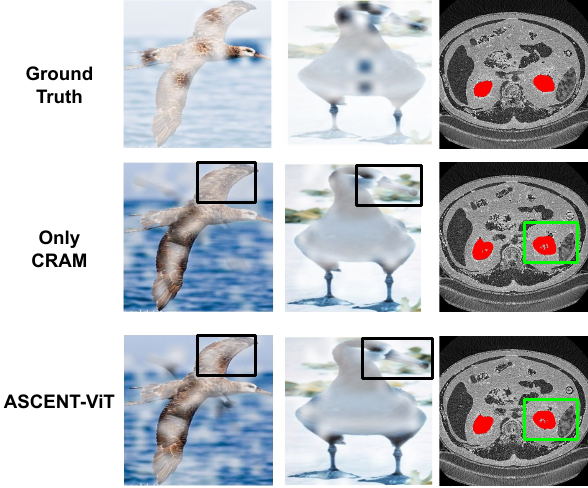}
    \caption{Visualizing concept attention scores over two samples from test set of CUB200 and one sample from KITS. We highlight the area of interest with a bounding box. (LEFT) ASCENT-ViT captures the brown front and back edge of wings better than utilizing only CRAM (i.e., assigns higher attention weights). (CENTER) ASCENT-ViT captures the annotations over the beak (tip) and eyes better than CRAM. (RIGHT) ASCENT-ViT captures the cysts annotation much more accurately as CRAM ignores some inner patches.}
    \label{fig:explain-img}
\end{figure}

\vskip -5pt
\subsection{Discussion}
Figure~\ref{fig:explain-img} shows two examples of correctly classified images from CUB200 and one from KITS. The first row (Ground Truth) represents the ground-truth human annotations of the concepts. The second and third rows represent the explanations produced using models trained using only CRAM and ASCENT-ViT on the ViT-Base backbone, respectively. We observe that ASCENT-ViT can better capture regions `missed' by only CRAM - the brown edge of wings in the first sample and beak shape in the second as compared to ground truth. All the pixels of the cysts in the kidney image are better captured by ASCENT-ViT as compared to only CRAM which misses relevant pixels in the middle of the cyst.

\begin{figure}
    \centering
    \includegraphics[width=0.4\textwidth]{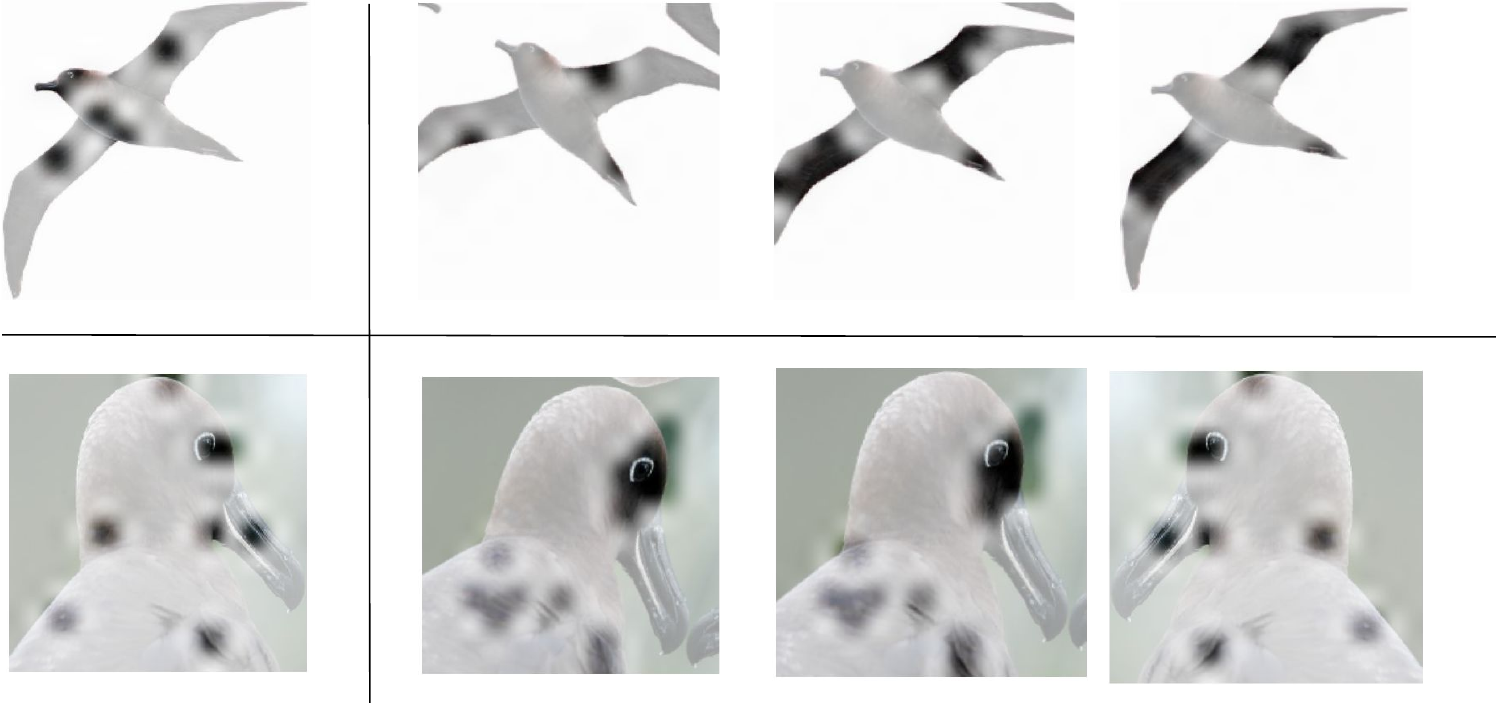}
    \caption{Robustness of concept explanations generated by ASCENT-ViT with transformations on test images as compared to ground truth (LEFT). Row-1 shows random rotations and crops across the center. Row-2 shows random flips and rotations.}
    \vskip -10pt
    \label{fig:robustness-transform}
\end{figure}

\noindent\textbf{Robustness to transformations.} We also report the qualitative performance of ASCENT-ViT in small transformations in Figure~\ref{fig:robustness-transform} - rotation, random cropping, and zoom where we observe that ASCENT-ViT is robust to minor transformations.

\noindent\textbf{Interventions.} Lastly, we show that ASCENT-ViT supports interventions better than CBMs and Only CRAM in case of wrong prediction (in Appendix).

\section{Conclusion}
\label{sec:conclusion}
In this paper, we propose Attention-based Scale-aware Concept Learning Framework for Enhanced Alignment in Vision Transformers (ASCENT-ViT), a concept-based explainability framework that aligns concepts with representations. ASCENT-ViT captures concepts across multiple scales using a Multi-scale Encoding Module and composes effective multiscale concept-patch relationships through a Multi-scale Feature Interaction Module. Finally, composite representations are aligned with concepts using a Concept-Representation Alignment Module. Quantitative and qualitative results demonstrate superior performance over existing methods on multiple ViT backbones. We hope our work helps in designing robust concept-based explainability modules for large-scale DNNs.

\bibliographystyle{named}
\bibliography{ijcai25}

\clearpage
\clearpage

\appendix
\appendix
\section{Appendix}

The Appendix is organized as follows:
\begin{itemize}
    \item Implementation Details
    \item Effect of Model Type
    \item Overhead of explainability modules
    \item Visual comparison across ViT architectures
    \item Robustness Analysis
    \item Additional Results on Pascal dataset
    \item Additional Results using ASCENT-ViT
    \item Intervention on Test Datasets
\end{itemize}

\subsection{Implementation Details}
\noindent \textbf{ViT Backbone.} 
We set the patch size to correspond to 16x16 pixels. Each image is resized to (224,224), hence a patch sequence is (224/16, 224/16) = (14,14). We further flatten patches and utilize pre-trained positional embeddings. Following \citet{rigotti2021attention}, we append the \textless CLS\textgreater token to the patch sequence after positional embeddings as it is designed to encapsulate global semantics. This results in an input sequence of length of 196+1 = 197 tokens. The internal embedding size ($dim$) of ViT is 1024.

\noindent \textbf{MSE Module.} We utilize three different scales in the Multi-scale Encoding Module, i.e., $S=3$. We utilize the scales - 1/8, 1/16 and 1/32. This corresponds to the scale-aware vector $\mathbf{c}$ composed of sizes: $c_1=1024$, $c_2=196$ and $c_3=16$, making the size of $\mathbf{c}$ be 1029. The first convolution block consists of three convolution blocks with the batch norm and ReLU activations followed by the Max pooling operation. The following convolution blocks consist of a single convolution layer of kernel size=3 and stride=2, as well as the batch norm and ReLU activations.  

\noindent \textbf{DMSF Module.} We utilize Multi-scale Deformable Attention \citep{zhu2020deformable}. We utilize 16 attention heads and four reference points, along with layer norms for each key, query, and value vector. Note that the output from the ViT backbone is appended with the \textless CLS\textgreater token. We strip the \textless CLS\textgreater token before passing through the DMSF module and append it to the output. The initialization value of $\mathbf{I}$ is set as 0.01. The value of $\psi$ is tunable and set as 1 for CUB, 2 for Concept-MNIST and 0.5 for AWA2.

\noindent \textbf{CRAM.} We utilize the same embedding dimensions, i.e., $dim=1024$ for Key, Query and Value vectors. The number of attention heads is set as 2 for both global and spatial concept attention matrices.

\noindent \textbf{Training Details.} We train each dataset and backbone for 50 epochs with early stopping. The maximum learning rate is set at 5e-5 with a linear warmup for the first 10 epochs followed by Cosine decay. Note that ViT training is negatively affected by state-less optimizers - as a consequence we use AdamW with 1e-3 weight decay. The weight of the explanation loss is set at 1.0. The batch size is 16 for each dataset with mixed precision (16-bit default) optimization.

\subsection{Effect of Model Type}
Note that CRAM in our method is similar to \citet{rigotti2021attention}. However, we amend the module to work for various ViT architectures. The performance on the models compared in \citet{rigotti2021attention} are on-par.

\noindent \textbf{Types of inductive bias.}  In this section, we first discuss the types of inductive biases in various ViT architectures. Inductive bias plays an important role in determining the characteristics of the learned embeddings from the ViT architecture. Note that the standard ViT architecture only properly encodes intra-patch relationships which can be thought of as a positional-only inductive bias. SWIN on the other hand, significantly differs from standard ViTs and only contains CNN-like operations through a shifted window implementation - making the inductive biases equivalent to CNNs. Finally, in addition to the primary inductive biases, self-supervised learning techniques also encode a transformation invariance inductive bias as in Dino. DeIT [4] is a similar architecture to ViT with a data-efficient training methodology that also encodes positional inductive biases.
We compare the performance against two CNN-only approaches, ST-CNN and MA-CNN \cite{rigotti2021attention} as well in Table~\ref{tab:model-type-effect-appendix} on the CUB200 dataset. We observe ASCENT-ViT outperforms all architectures.

\textbf{NOTE:} SWIN (1x1) is an improved version of the SWIN baseline where the final patch embedding is passed through a 1x1 convolution layer. We utilize the updated version in all the experiments in the paper.

\begin{table}[h]
    \centering
    \resizebox{0.49\textwidth}{!}{
    \begin{tabular}{c|c|c|c}
        \hline
         \textbf{Model Name} & \textbf{Inductive Bias} & \textbf{Only CRAM} & \textbf{ASCENT-ViT} \\
         \hline
         ST-CNN & CNN-like &  82.01 $\pm$ 0.01 & \textbf{82.56 $\pm$ 0.01} \\
            MA-CNN & CNN-like  & 82.84 $\pm$ 0.03 & \textbf{83.32 $\pm$ 0.01} \\
            \hline
         ViT-Large & Positional (Pos.) & 86.31 $\pm$ 0.2  & \textbf{87.26 $\pm$ 0.3}  \\
         DeIT-Base & Positional (Pos.) & 76.81 $\pm$ 0.8 & \textbf{76.96 $\pm$ 0.3}  \\
         SWIN (1x1) & CNN-like & 86.44 $\pm$ 0.8 & \textbf{ 86.72 $\pm$ 0.6} \\
         ViT-Base (Dino) & Pos. + Invariance & 76.5 $\pm$ 0.2 & \textbf{77.2 $\pm$ 0.1} \\
         \hline
    \end{tabular}}
    \caption{Performance on various ViT architectures on the CUB200 dataset using only CRAM and ASCENT-ViT.}
    \label{tab:model-type-effect-appendix}
\end{table}

\subsection{Overhead of Explanations}
Table~\ref{tab:model-param-effect} lists the percentage of parameters in the standard Feed Forward Network (FFN), CRAM, and ASCENT-ViT when used as the classifier head as a percentage of total model parameters. CRAM utilizes a modestly higher number of parameters as compared to FFN while ASCENT-ViT (with the MSE and DMSF modules) introduces slightly more parameters than CRAM. Overall, the number of additional parameters remains under 3\% of total model parameters. We can conclude that both ASCENT-ViT and CRAM provide good explainability-efficiency tradeoffs.

\begin{table}[h]
    \centering
    \resizebox{0.49\textwidth}{!}{
    \begin{tabular}{c|c|c|c|c}
        \hline
         \textbf{Model Name} & \textbf{\# Params} & \textbf{FFN} &  \textbf{CRAM} & \textbf{ASCENT-ViT} \\
         \hline
         ViT-Base & 86M &  $<$1\% & 2\% & 2.4\% \\
         ViT-Large & 307M &  $<$1\% & $\leq$1\% & $\leq$1\% \\
         DeIT-Base & 86M & $<$1\% & 2\% & 2.5\% \\
         SWIN  & 88M &  $<$1\% & 2\% & 2.2\% \\
         ViT-Base (Dino) & 86M &  $<$1\% & 2\% & 2.4\% \\
         \hline
    \end{tabular}}
        \caption{Percentage of additional parameters introduced by specific classifier heads of various ViT architectures. CRAM and ASCENT-ViT offer improved explainability at a fraction of the additional parameter cost.}
    \label{tab:model-param-effect}
\end{table}

\begin{figure*}[h]
    \centering
    \includegraphics[width=0.6\textwidth]{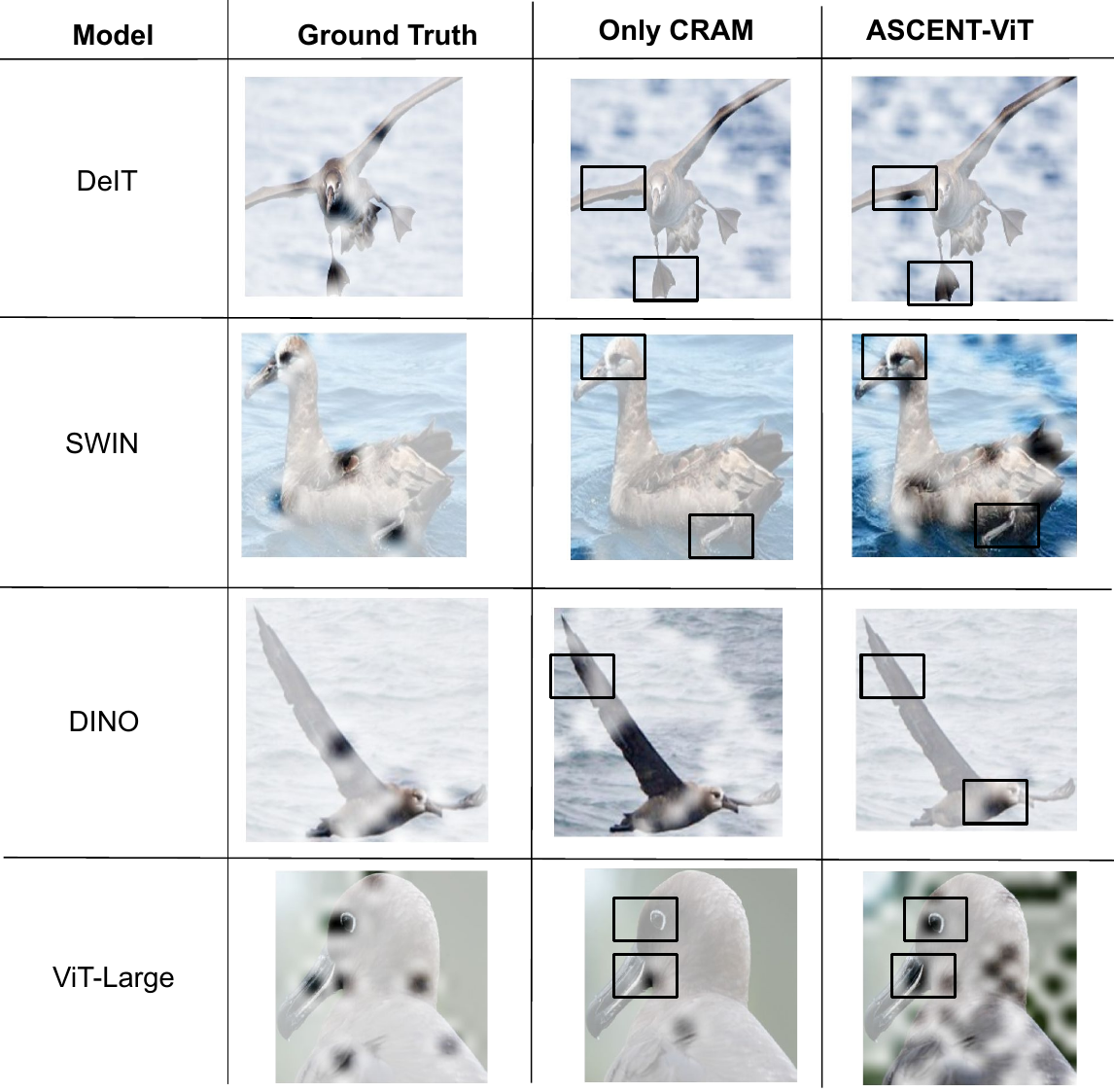}
    \caption{Comparision of explanation visualizations across different ViT architectures. The first column denotes the model architecture. Selected samples from the test set correctly classified by the model are represented in the second, third and fourth columns. The second column represents manually annotated ground truth (color) overlayed on the image. The third and fourth columns show the attention map of the CRAM and ASCENT-ViT methods. The bounding boxes are areas of interest where ASCENT-ViT captures detailed explanations not captured by CRAM.}
    \vskip -10pt
    \label{fig:viz-compare-model}
\end{figure*}

\subsection{Visual Comparison across ViT Architectures}
Figure~\ref{fig:viz-compare-model} shows some correctly classified examples from the CUB200 dataset. We compare random test set images across the four ViT architectures overlayed by the attention explanations. We observe that CRAM fails to capture sufficient concept annotations (compared to ground truth) while ASCENT-ViT captures the concept annotations well. The bounding boxes (in black) are areas of interest where ASCENT-ViT outperforms CRAM. Note that we utilize the same setting for background suppression as in \citet{rigotti2021attention} for a fair comparison - which results in a few additional background pixels being identified as important.

\subsection{Robustness Analysis}
In addition, we demonstrate the robustness of our approach to transformations applied to the test images in Figure~\ref{fig:robustness-transform} (Main Text). The images in the first column are the ground-truth concept annotations. The first row shows the concept explanations identified on the images with rotation and crop transformations. We observe that ASCENT-ViT can correctly identify the front and back portions of the wing as concepts. Similarly, Row-2 shows concepts identified under rotation and flip transformations. Once again ASCENT-ViT can identify the correct concept annotations like the eyes and beak areas.

\subsection{Additional Results on Pascal Dataset}
We report additional results on the Pascal dataset - concept loss and prediction performance over different values of $\psi$. Note that Pascal dataset has crude annotations of concepts making concept-learning noisy. Table~\ref{tab:pascal} shows the effect of a feed-forward network (FFN), CRAM and ASCENT-ViT. Note that the backbone ViT architecture utilized is the ViT-Large and concept errors are the MSE loss. We observe that both CRAM and ASCENT-ViT perform on par with both prediction and concept errors.

\begin{table}[h]
    \centering
    \resizebox{0.5\textwidth}{!}{
    \begin{tabular}{c|c|c|c}
     \hline
    \textbf{Class. Head} & \textbf{Spatial Wt. ($\psi$)} & \textbf{Concept Error} & \textbf{Accuracy (\%)}  \\
      \hline
    FFN & - & - & 81.3  \\
    \hline
    CRAM & 0 & 0.089 & 81.9  \\ 
     \hline
    \multirow{3}{*}{ASCENT-ViT} & 0.1 & 0.091 & 80.4  \\
    & 1 & 0.088 & \textbf{81.9} \\
    & 10 & \textbf{0.083} & 81.7 \\
     \hline
    \end{tabular}}
    \caption{Results on the Pascal dataset with classifier heads as a feed-forward network (FFN), CRAM, and ASCENT-ViT on the ViT-Large backbone. We observe that ASCENT-ViT and CRAM perform on par.}
    \label{tab:pascal}
\end{table}

\begin{figure}[h]
    \centering
    \includegraphics[width=0.45\textwidth]{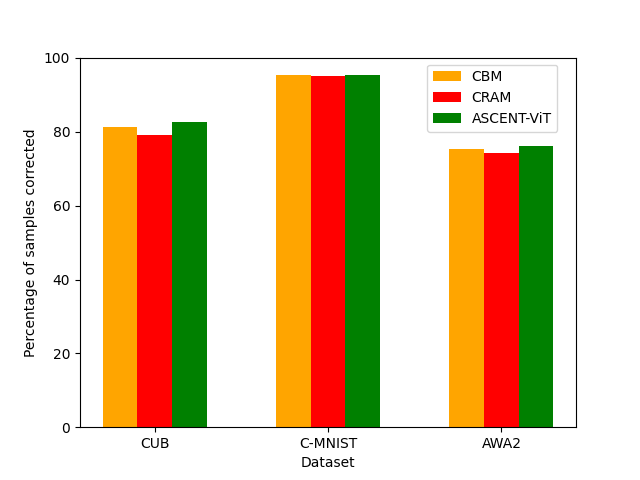}
    \caption{Test-time Intervention on datasets - CUB, MNIST (prediction) and AWA2 using CBMs (Orange), CRAM (Red) and ASCENT-ViT (Green).}
    \label{fig:intervene}
\end{figure}
\begin{figure}[h]
    \centering
    \includegraphics[width=0.48\textwidth]{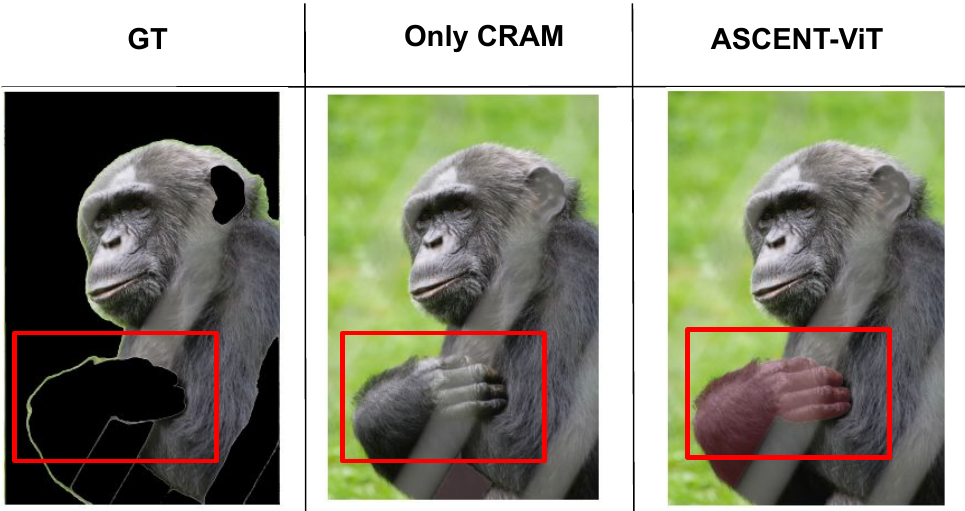}
    \caption{A few more results using ASCENT-ViT on a randomly chosen, correctly classified test image from the AWA2 dataset along with the correctly classified concepts. The concept "ARM" in the first (LEFT) image has been computed using SAM model. The second and third images show the concept captured by only CRAM and ASCENT-ViT. As can be seen ASCENT-ViT captures the ARM concept while only CRAM does not. }
    \label{fig:acam-3}
\end{figure} 


\subsection{Intervention on Test Datasets}
The most important property of a human-in-the-loop system is the ability to intervene and `correct' the incorrect predictions. We demonstrate the interventions on the concepts for all datasets - MNIST (prediction), CUB, and AWA2 for CRAM and ASCENT-ViT (Figure~\ref{fig:intervene}). We compare the success rate with CBMs. The percentage of incorrectly classified samples corrected after intervention is represented in the y-axis. We observe that intervening on ASCENT-ViT module beats both CRAM and CBM. Note that we can only intervene on GLobal Concepts for all datasets, i.e., $C_{Global}$. Surprisingly, the CRAM in itself underperforms on CBMs.

\subsection{Additional Visual Results using ASCENT-ViT} 
\begin{figure}[h]
    \centering
    \includegraphics[width=0.49\textwidth]{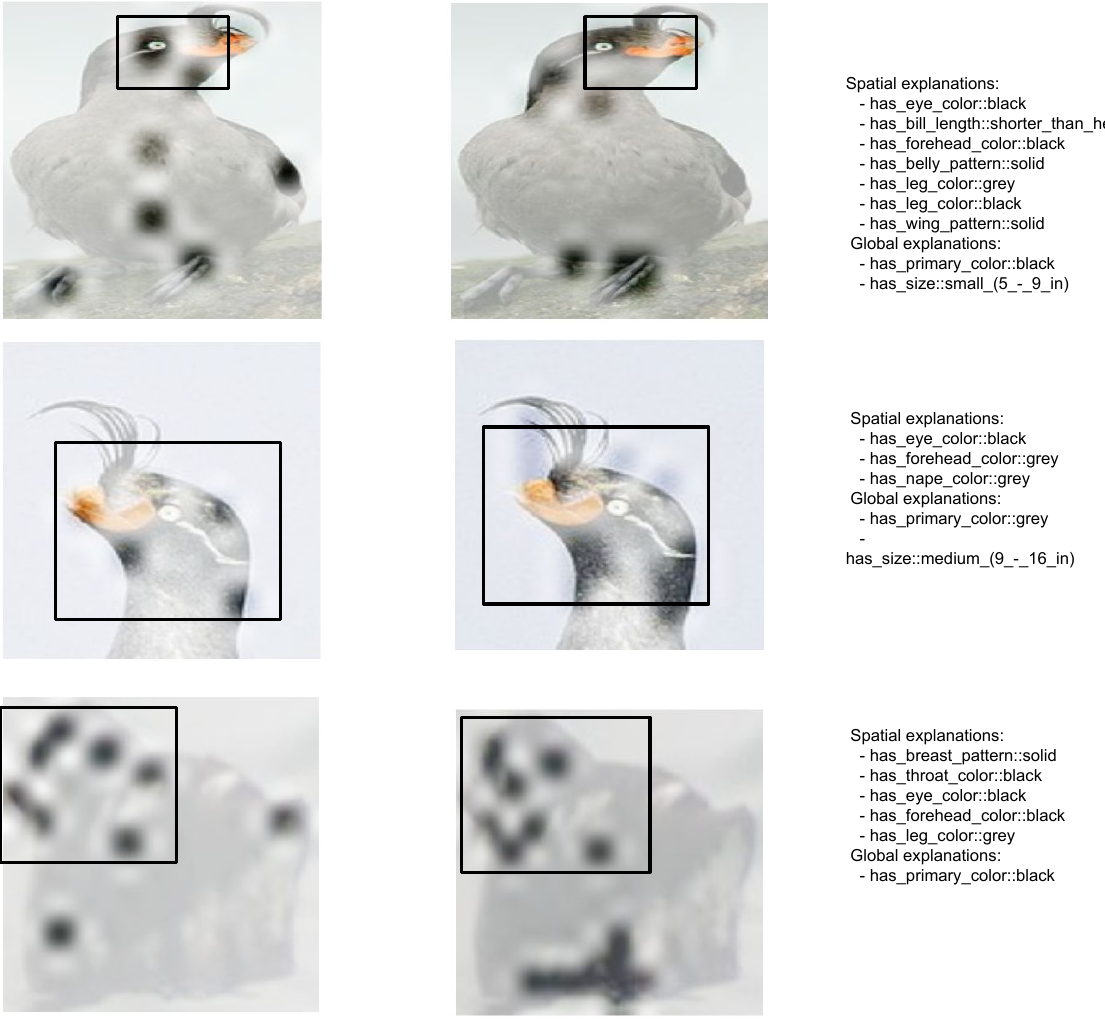}
    \caption{A few more results using ASCENT-ViT on 3 randomly chosen, correctly classified test images from the CUB200 dataset along with the correctly classified concepts.}
    \label{fig:additional}
\end{figure}

\begin{figure}
    \centering
    \includegraphics[width=0.49\textwidth]{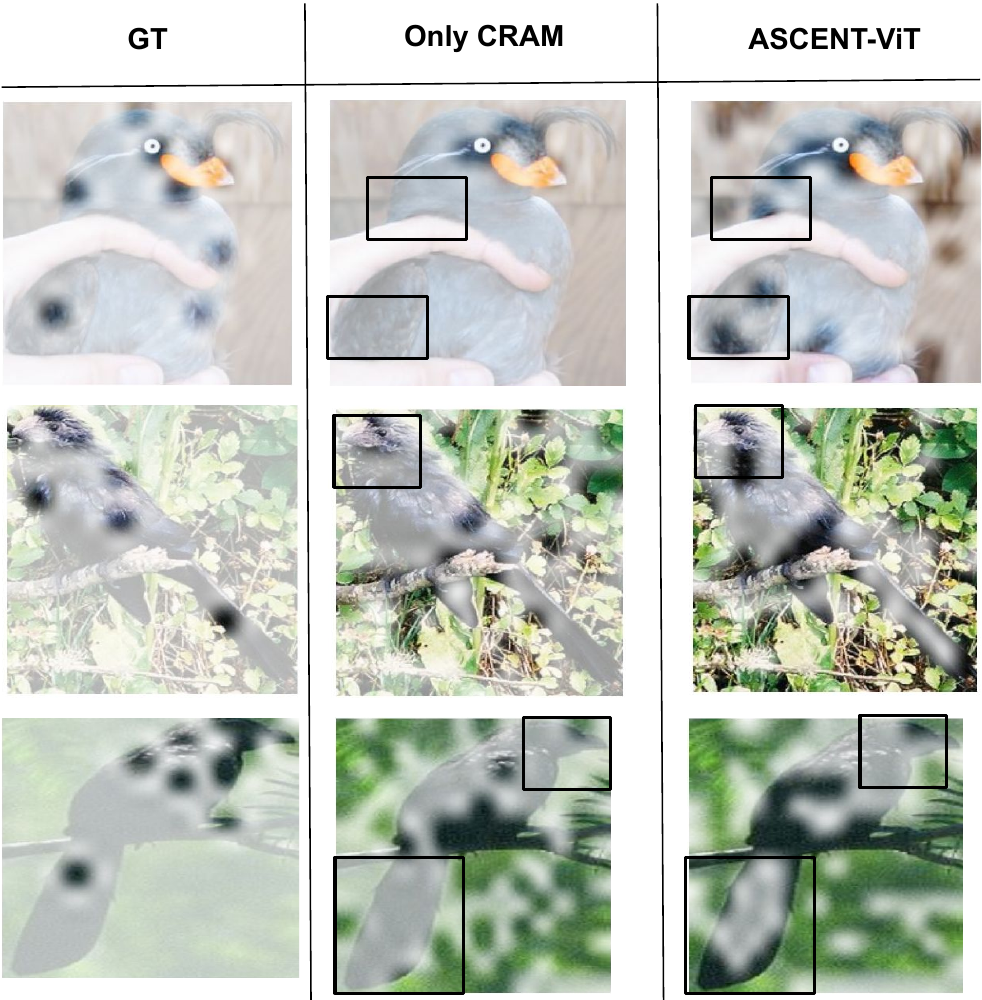}
    \caption{A few more results comparing CRAM and ASCENT-ViT generated concept explanations on 3 randomly chosen, wrongly classified test images from the CUB200 dataset. Note that even when the images are classified incorrectly, ASCENT-ViT learns more accurate concept maps. In all the images, we mark the area of interest through a black bounding box. In Row-1, ASCENT-ViT is successfully able to capture the Nape and Wing concepts missed by only the CRAM module. In Row-2, ASCENT-ViT identifies the head and under-eyes concepts better than CRAM. Finally, in Row-3, ASCENT-ViT identifies the head and tail concepts missed by CRAM. }
    \label{fig:additional-acam-cpa}
\end{figure}

Figure~\ref{fig:acam-3} shows a correctly classified sample from the AWA2 dataset. Figure~\ref{fig:additional} demonstrate additional results as compared to ground-truth concept annotations for correctly from the CUB200 dataset.  Figure~\ref{fig:additional-acam-cpa} incorrectly classified samples. For the correctly classified samples in Figure~\ref{fig:additional} report the verbose concepts subdivided into spatial and global categories. 

\end{document}